\documentclass[conference]{IEEEtran}
\IEEEoverridecommandlockouts
\usepackage{cite}
\usepackage{amsmath,amssymb,amsfonts}
\usepackage{algorithmic}
\usepackage{graphicx}
\usepackage{textcomp}
\usepackage{xcolor}
\usepackage{url}
\def\BibTeX{{\rm B\kern-.05em{\sc i\kern-.025em b}\kern-.08em
    T\kern-.1667em\lower.7ex\hbox{E}\kern-.125emX}}
\begin{document}

\title{Early Stage Diabetes Prediction via Extreme Learning Machine\\
}

\author{\IEEEauthorblockN{Nelly Elsayed}
\IEEEauthorblockA{\textit{School of Information Technology} \\
\textit{University of Cincinnati}\\
Cincinnati, Ohio, United States \\
nelly.elsayed@uc.edu}
\and
\IEEEauthorblockN{Zag ElSayed}
\IEEEauthorblockA{\textit{School of Information Technology} \\
\textit{University of Cincinnati}\\
Cincinnati, Ohio, United States \\
elsayezs@ucmail.uc.edu}
\and
\IEEEauthorblockN{Murat Ozer}
\IEEEauthorblockA{\textit{School of Information Technolog} \\
\textit{University of Cincinnati}\\
Cincinnati, Ohio, United States\\
ozermm@ucmail.uc.edu}
}

\maketitle

\begin{abstract}
Diabetes is one of the chronic diseases that has been discovered for decades. However, several cases are diagnosed in their late stages. Every one in eleven of the world's adult population has diabetes. Forty-six percent of people with diabetes have not been diagnosed. Diabetes can develop several other severe diseases that can lead to patient death. Developing and rural areas suffer the most due to the limited medical providers and financial situations. This paper proposed a novel approach based on an extreme learning machine for diabetes prediction based on a data questionnaire that can early alert the users to seek medical assistance and prevent late diagnoses and severe illness development. 

\end{abstract}


\section{Introduction}
\begin{figure}[t]
    \centering
	\includegraphics[width= 8.8cm, height=3.5cm]{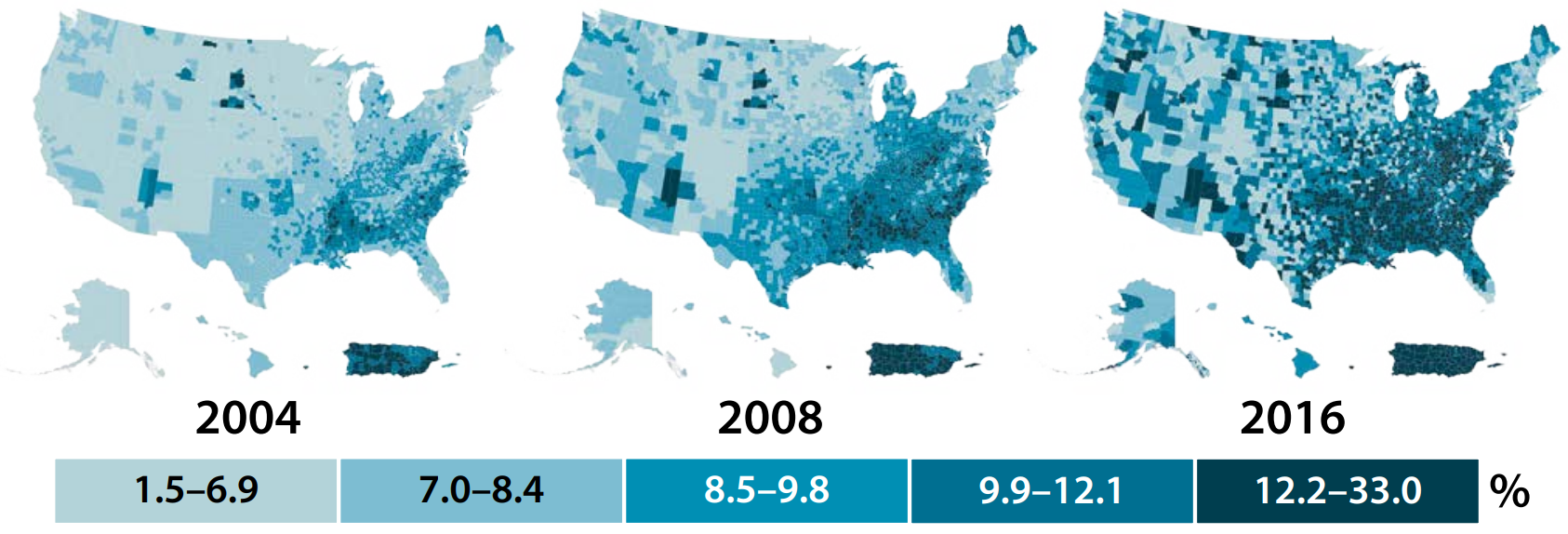}
	\caption{County-level prevalence of diagnosed diabetes among adults aged 20 years or older,
United States, 2004, 2008, and 2016~\cite{diabetesRateReport}.}
	\label{diabetesRateG}
\end{figure}

\begin{figure*}
	\centering
	\includegraphics[width= 18 cm, height=8.5cm]{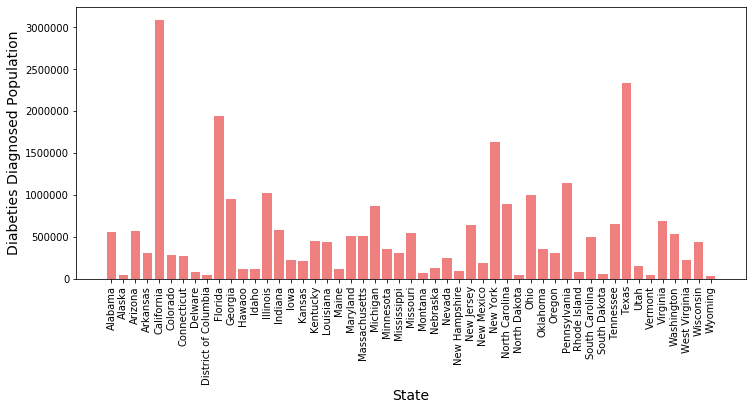}
	\caption{The number of adults that have been diagnosed by diabetes in each state at US~\cite{american_diab_report}.}
	\label{diab_by_state}
\end{figure*}

Diabetes is one of the diseases that has aggressively spread the world for several decades. Diabetes is a metabolic chronic disease caused by the malfunction of the pancreas production of the insulin hormone. This is characterized by elevated levels of blood sugar (or blood glucose) that the insulin hormone responsible for entering it to the cells to produce energy~\cite{mayo_definition,who_diab}. One in eleven of the world's adult population is suffering from diabetes. 46\% of people with diabetes are undiagnosed. According to the American Diabetes Association, Figure~\ref{diab_by_state} shows the number of adults that have been diagnosed with diabetes in each state in the United States~\cite{american_diab_report}. Untreated and uncontrolled diabetes leads over time to severe damage to body organs, especially blood vessels, brain, heart, eye, nerves, and kidneys~\cite{who_diab}. Besides, it can lead to strokes and sudden death~\cite{mayo_definition,american2002screening}. Figure~\ref{diabetesRateG} shows the national diabetes diagnosis rate over the past 16 years.

Diabetes can be categorized into one of the following types: type 1 diabetes, type 2 diabetes~\cite{diab_types}, gestational diabetes, pre-diabetes, monogenic diabetes~\cite{monogenic}, and cystic fibrosis-related diabetes~\cite{Cystic}.

Type 1 diabetes is a chronic condition in which the pancreas produces little or no insulin, which is thought to be caused by an autoimmune system reaction that destroys beta cells; however, The exact cause of type 1 diabetes is unknow~\cite{atkinson2014type,devendra2004type}. Insulin is a hormone needed to allow glucose to enter cells to produce energy. Glucose comes from food and liver stored exceed glucose as glycogen. When the glucose levels are low, such as when a person fasten, the liver breaks down the stored glycogen into glucose to keep the person's glucose levels within a normal range. Type 1 diabetes is usually diagnosed in children between four and seven years old and in children between 10 and 14 years old. Approximately 5-10 percent of people with diabetes have type 1~\cite{cdc_diab1}.
In type 1 diabetes (T1D), glucose does not reach cells, causing a high concentration of sugar in the blood, and the body cells starve. Over time, T1D can affect major organs in a person's body, including the heart, blood vessels, nerves, and eyes. T1D can dramatically increase the risk of heart and blood vessel disease, nerve damage (neuropathy), kidney damage (nephropathy), foot damage, skin conditions, and pregnancy complications.

\begin{figure}[h]
    \centering
	\includegraphics[width= 6.5cm, height=12cm]{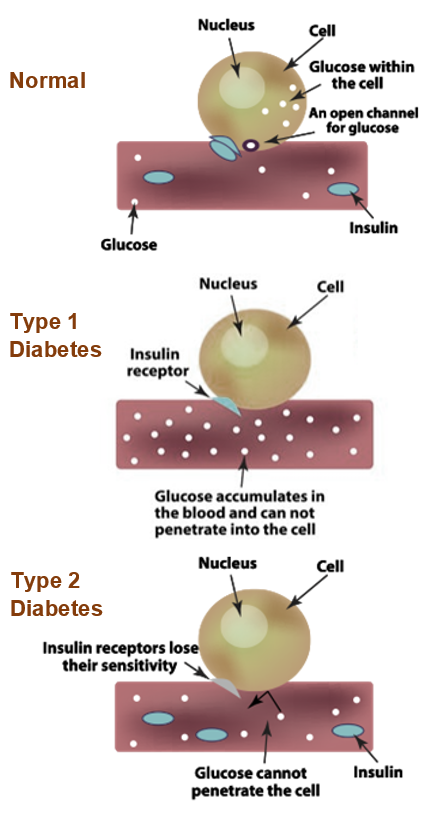}
	\caption{Comparison between Normal, Type 1 Diabetes and Type 2 Diabetes~\cite{InToLife}.}
	\label{Type1and2}
\end{figure}  

Type 2 diabetes (T2D) is a chronic condition that affects the way your body uses insulin the way it should be (Figure~\ref{Type1and2}). Usually, T2D is known as adult-onset diabetes that may occur due to previous obesity where the body becomes resistant to insulin or when the pancreas is unable to produce enough insulin~\cite{chatterjee2017type,olokoba2012type}. It is the most common type of diabetes. There are about 29 million people in the U.S. with type 2. Another 84 million have pre-diabetes~\cite{cdc_diab2}.

Long-term complications of diabetes develop gradually; however, they can eventually be disabling or even life-threatening because high blood sugar can damage and cause problems with Heart and blood vessels, kidneys, nerves, skin, sleep, brain at higher risk of Alzheimer's disease and depression. Figure~\ref{diabetesRateG} shows the rate of diabetes diagnoses.

The other types of diabetes could be curable or temporary. For example, Gestational Diabetes Mellitus (GDM) is a type of diabetes first seen in a pregnant woman with no preexisting diabetes condition before pregnancy. GMD usually shows up in the middle of pregnancy (between 24 and 28 weeks)~\cite{gestationalDiabetes,american2004gestational}. GMD generates problems for the mother and the baby, including an extra-large baby, cesarean section surgery, Preeclampsia, Hypoglycemia, polyhydramnios, macrosomia, organomegaly, fetal hypoxia respiratory distress syndrome in the newborn, hypoglycemia, polycythemia, hyperbilirubinemia, and the risk of developing T2D. Therefore, early diagnosis is essential to indicate adequate medical follow-up and treatment in a timely manner~\cite{gdmOutcomes}.

Prediabetes is a health condition in which blood sugar levels are higher than normal but not high enough to be diagnosed as type 2 diabetes. More than one in three have Americans have prediabetes where \textgreater84\% do not know they have it. Prediabetes increases the risk of developing type 2 diabetes, heart disease, and stroke. However, prediabetes can be cured and prevented early via making lifestyle changes to prevent or delay the complications~\cite{prediabetes}. Thus early prediction and high sensitivity accurate diagnosis are critical.

Monogenic diabetes (MD) is a rare form that combines the characteristics of both T1D and T2D. MD is often misdiagnosed as it results from mutations or changes in a single gene and is called monogenic~\cite{steck2011review}. In counts for 1\% to 4\% of all cases of diabetes in the United States~\cite{monogenicDiabetes}, MD appears in several forms and most often affects young people, mainly \textless25 years. In most forms of the disease, the body is less able to make insulin. MC can be in the founding of two forms: a Maturity-Onset Diabetes of the Young (MODY)~\cite{shields2010maturity}, which is the most common form, or Neonatal diabetes that is usually in infants (\textless6 months old) can be permanent of a transient.
Genetic testing can diagnose most forms of monogenic diabetes starting as early as six months old. A correct diagnosis with proper treatment should lead to better glucose control and improved health in the long term~\cite{monogenicGeneticTest}.

Cystic fibrosis-related diabetes (CFRD) is caused by thick, sticky mucus that is characteristic of the disease that causes scarring of the pancreas\cite{moran2009cystic}. This scarring prevents the pancreas from producing normal amounts of insulin-making the illusion "insulin resistant." CFRD is the most common disorder in Caucasian populations affecting ($\sim$70,000) globally~\cite{CFRDStatistics}. Some CFRD patients may not experience any symptoms. However, many people with CFRD do not know they have CFRD until they are tested for diabetes. Treatment may be effective at an earlier stage to prevent deterioration of lung function~\cite{CFRDTreatment}.


Due to life commitments, financial constraints, and the limited number of medical providers, especially in rural and development areas, populations are not seeking regular health checkups that lead to late diagnoses of diseases that may cause severe illness due to untreated or late treatment~\cite{cook2008ethics,merwin2006differential}. Moreover, due to the COVID-19 pandemic, the medical providers are overwhelmed with the COVID-19 cases, which postpone several non-urgent medical procedures and visits, which may delay the detection of several diseases, especially diabetes. 

In this paper, we proposed a novel model to predict diabetes from a questionnaire that users can use, which raises a health alarm to seek medical assistance and check diabetes. This model is inexpensive from the implementation and user aspects, and the model can even be used by medical personnel as an assistant for finding diabetes. We used the extreme learning machine (ELM) as a rapid and cost inexpensive model to implement and design as well as it showed a significant accuracy. The model has been employed within a user-friendly application to simplify the application use. Therefore, the model could be efficiently used on various data sources, which increases the usability of the proposed model.

\section{Extreme Learning Machine (ELM)}

The Single-Hidden-Layer Feedforward Network (SLFN) is the simplest type of feedforward network which architecture design consists of a single-hidden-layer perceptron~\cite{maan2016survey}.
The first Extreme Learning Machine (ELM) was proposed by Huang et al.~\cite{huang2006extreme} as a simple learning algorithm for a Single Hidden Layer feedforward neural network (SLFN) with $N$ hidden nodes~\cite{matias2014learning,hornik1991approximation}. The inputs to the SLFN are directly fit into the outputs through a single series of a set of weights. The sum of weights and input products are calculated in each node. The activation function of the SLFN neurons uses the linear threshold unit principles~\cite{maan2016survey}. The SLFN can differentiate between two to $N$ observations using any non-linear activation functions such as hyperbolic tangent (tanh) or hard limit (hardlim). Nevertheless, it mainly depends on the number of neurons within the single hidden layer. When the SLFN consists of $N$ hidden neurons, it can precisely differentiate between $N$ distinct observations~\cite{huang1998upper,huang2000classification}. However, the main issue of the SLFN is the generalization problems which is a significant drawback of creating such a model using traditional learning algorithms.

During the training process, the ELM randomly selects the hidden layer nodes. Therefore, the ELM algorithm revs the learning process and obtains a better generalization performance than the other traditional feedforward network learning algorithms~\cite{huang2006extreme}. 

\begin{figure}
	\centering
	\includegraphics[width=7.5cm,height=2.1cm]{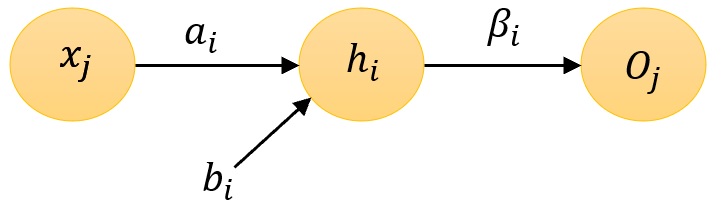}
	\caption{The network architecture of a single hidden layer of ELM network.}
	\label{elm}
\end{figure}

Figure~\ref{elm} shows the network architecture of a single hidden layer of ELM network ith $L$ neurons. $\mathrm{x_j}$ is the $j^{th}$ input vector of the training sample $\textbf{x} $ of size $N$ and $O_j$ is the $j^{th}$ output of the ELM hidden neuron. The output weight of the $i^{th}$ hidden neuron (node) is denoted by $\beta_i$. $a_i$ and $b_i$ are the input weight vector and the bias vector for the hidden neuron $i$, respectively. Accordingly, the output function of the ELM can be calculated as follows:
{\small
\begin{equation}
f_L(\textbf{x}) = \sum_{i=1}^{L} \beta_i h_i (\textbf{x})
\end{equation}
}
\noindent
where $h_i(\textbf{x})$ is the output function of the $i^{th}$ hidden node.

The ELM has shown a significant success in different applications such as text categorization~\cite{zheng2013text}, classification and arrhythmia detection of electrocardiogram signals~\cite{karpagachelvi2012classification,elsayed2020simple}, image segmentation~\cite{pan2012leukocyte}, and regression problems~\cite{feng2012evolutionary}. The ELM algorithm's main characteristics tend to find the smallest norm of weights that leads to a robust generalization performance of the network~\cite{bartlett1998thesamplecomplexityofp,huang2006extreme} in addition to the tendency to find the smallest training error, which only focuses on the traditional feedforward network learning algorithms~\cite{huang2006extreme}. Moreover, the ELM acts like a sort of neural network regularization but with non-tuned hidden layer mappings~\cite{huang2015extreme}. Thus, the ELM has a smaller overall training error, robust generalization performance, and a faster learning process than traditional multilayered feedforward neural networks. In addition, during our empirical design of the detection model.

\section{Proposed Model}\label{model_arch}
\begin{figure}
	\centering
	\includegraphics[width=8cm,height=2.5cm]{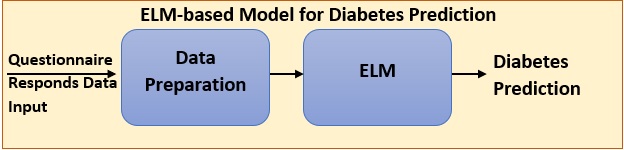}
	\caption{The proposed ELM-based model architecture for diabetes prediction.}
	\label{elm-d}
\end{figure}

The proposed ELM-based model is designed and implemented to predict diabetes based on questionnaire responses. We intentionally developed the ELM-based model as a simple model with high detection performance to accelerate the learning process and reduce the hardware processing requirements. Therefore, the model is cheaper to implement, and it helps to reduce the CO2 footprint~\cite{baumann2017co2}. The model architecture of the proposed ELM-based model for diabetes prediction is shown in Figure~\ref{elm-d}. The model consists of two main stages: data preparation and the ELM-learning component. In the data preparation stage, the questionnaire responses are normalized, and the text responses are converted to the corresponding numeric values. Then, the data after the preprocessing is used to train the ELM model to predict diabetes. Finally, the ELM-based model can successfully predict whether the patient has developed diabetes or its absence.  
Then, the model is applied within a user-friendly application where the users can answer the set of health-related questions to predict whether the user has developed diabetes or not. This application can play a significant role in helping prevent developing severe illnesses due to late diabetes detection. In addition, it is a non-expensive model which can be implemented using a low hardware budget. The model does not replace medical screening; however, it can be significant for use in rural and developing countries where many people can not afford regular medical checks to provide an alert to the patients to seek professional medical assistance.

\subsection{Experiments and results}

\section{Dataset}
We designed our model using the Early Stage Diabetes Risk Prediction dataset benchmark. The data set is available in the IEEE Dataport~\cite{k01r-x481-21} and the  UCI Machine Learning Repository~\cite{misc_early_stage_diabetes_risk_prediction_dataset._529}. The dataset is a multivariate dataset that contains the sign and symptom data of newly diabetic or would-be diabetic patients. The dataset has been collected using direct questionnaires from the patients of Sylhet Diabetes Hospital in Sylhet, Bangladesh, and approved by a doctor. The participants of the dataset questionnaire represent male and female genders and the age range between 20 to 65 years old. The dataset consists of 17 attributes representing different health-related questions and the class attribute (diabetes or non-diabetes). The dataset questionnaire attributes and their corresponding possible values are shown in Table~\ref{data_table}.

\begin{table}
	\caption{The Dataset Questionnaire Attributes.}
	\begin{center}
		\begin{tabular}{|l|l|}
			\hline
			\textbf{Attribute} &\textbf{Value}\\
			\hline
			Age & 20-65\\
			Gender& Male/Female\\
			Polyuria& Yes/No\\
			Polydipsia& Yes/No\\
			Sudden weight loss& Yes/No\\
			Weakness & Yes/No\\
			Polyphagia& Yes/No\\
			Genital thrush& Yes/No\\
			Visual blurring& Yes/No\\
			Itching&Yes/No\\
			Irritability&Yes/No\\
			Delayed healing&Yes/No\\
			Partial paresis&Yes/No\\
			Muscle stiffness&Yes/No\\
			Alopecia&Yes/No\\
			Obesity&Yes/No\\
			Class& Diabeties/Normal\\
			
			\hline
		\end{tabular}
		\label{data_table}
	\end{center}
\end{table}

During the data preparations stage, the age attribute has been normalized, the questions with yes or no possible responses were converted to binary values (zero and one) where zero represents no, and one represents yes. The class diabetes was represented as category one and the normal as category zero. Then, we split the dataset to train, validation, and test sets with ratios 70\%, 10\%, and 20\%, respectively.
\subsection{Model Setup and Results}

The model was implemented using Python 3.8, Sklearn 0.24.1, and Numpy 1.19.5 on an Intel(R) i7-9750H CPU @2.60GHz processor, 32 GB RAM, and 64-bit Microsoft Windows 10 operating system.
We used the extended Sklearn library to implement the ELM model. We set the number of hidden neurons to 50. The $\alpha$ mixing parameter was set to 1.0 to achieve a 100\% input activation. We used the RBF multiquadratic transfer function~\cite{pottman1992identification} which can be calculated by:
\begin{equation}
	\varphi(r)= e^{-(\epsilon r)^2}
\end{equation}
\noindent
where $\varphi : [0,\infty) \rightarrow R $ is the RBF multiquadratic activation function. We set the choice on the RBF multiquadratic function based on empirical evaluation of using different activations functions such as the hard limit (hardlim), the hyperbolic tangent (tanh), sine, triangular basis (tribas), and gaussian functions. The RBF multiquadratic activation function shows the highest accuracy for diabetes prediction with a minimal false-positive rate.

\begin{table}
	\caption{The Emperical Results of the ELM-based Model Using Different Transfer Functions.}
	\begin{center}
		\begin{tabular}{|l|c|c|c|c|c|}
			\hline
			\textbf{Transfer}&\multicolumn{5}{|c|}{\textbf{Metrics}} \\
			\cline{2-6} 
			\textbf{Function} & \textbf{\textit{Precision}}& \textbf{\textit{Recall}}& \textbf{\textit{F1-Score}}& \textbf{\textit{Accuracy}}&\textbf{\textit{Time}} \\
			\hline
			hardlim&0.80&0.82&0.81&83.65\%&0.164\\
			tanh&0.61&0.32&0.42&62.50\%&\textbf{0.125}\\
			sine&0.60&0.89&0.72&70.19\%&0.172\\
			tribas&0.58&0.84&0.69&67.30\%&0.172\\
			gaussian&0.79&0.77&0.78&81.73\%&0.187\\
			multiquadratic&\textbf{0.96}&\textbf{1.00} &\textbf{0.98} & \textbf{98.07\%}&0.157\\
			\hline
		\end{tabular}
		\label{accuracies}
	\end{center}
\end{table}

Table~\ref{accuracies} shows the prediction accuracy of the ELM model using different transfer functions. In addition, it shows the precision, recall, F1-score, and the train time (in seconds) where the precision, recall, and F1-score are calculated using \eqref{precision}, \eqref{recall}, and \eqref{f1score}, respectively. Table~\ref{accuracies} also shows the time efficiency of training the proposed ELM-based model regardless of the selected transfer function. 

	\begin{equation}
		precision = \frac{True Positive}{True Positive+False Positive}\\
		\label{precision}
		\end{equation}
	\begin{equation}
		recall= \frac{True Positive}{True Positive+False Negative}\\
		\label{recall}
	\end{equation}
	\begin{equation}
		\mathit{F1\textnormal{-}score} = 2 \times \frac{precision \times recall}{precision + recall}
		\label{f1score}
		\end{equation}
\begin{figure}
	\centering
	\includegraphics[width=6cm,height=5cm]{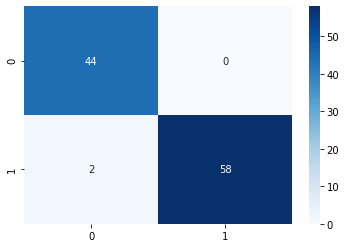}
	\caption{The confusion matrix of the ELM-based model for predicting diabeties applying the multiquadratic transfer function.}
	\label{confusion_matrix}
\end{figure}

The confusion matrix of the proposed EML-based model for predicting diabetes applying the multiquadratic transfer function is shown in~\ref{confusion_matrix}, where the label 0 indicates the normal and label 1 indicates diabetes. The model achieved zero false positive rates and only 1.9\% of the false negative rate. 

The proposed model showed a significant accuracy for predicting diabetes using non-expensive hardware support.

\section{Conclusion and Discussion}
Diabetes is a silent killer disease that attacks different people from different age categories and genders. Late detection of diabetes may occur in severe, long-term, life-threatening diseases. Due to the COVID-19 pandemic, there is a significant shortage of medical providers and long waiting lines for medical services. In addition, many people postpone their health checkups due to life commitments such as work and family, as well as due to financial-related aspects. Moreover, there is a shortage of professional labs in rural and development areas, and seeking a healthcare provider may require traveling, which requires extra cost. In this paper, we proposed a machine learning model that is implemented using the ELM network, which, based on health questionnaire data, can predict diabetes in adults. The model achieved 98.07\% accuracy with zero false-positive rates. In addition, the model can be easily implemented, and it is inexpensive from the implementation and hardware requirements aspect. The model does not replace the medical provider. However, it can provide an early alarm to the user to seek professional medical support, early detect diabetes, and receive the needed treatment. The proposed model will be employed within a user-friendly application that can be used in different regions and areas, especially the rural and development areas.


\section*{Acknowledgment}
This material is based upon work supported by the College of Education, Criminal Justice, Human Services and Information Technology, at the University of Cincinnati under FY 22 CECH Diversity Research Grant funding.

\bibliography{references}

\begin{thebibliography}{10}

\bibitem{mayo_definition}
M.~Clinic, ``Diabetes.''
  \url{https://www.mayoclinic.org/diseases-conditions/diabetes/symptoms-causes/syc-20371444},
  2020.

\bibitem{who_diab}
W.~H. Organization, ``Diabetes.''
  \url{https://www.who.int/health-topics/diabetes#tab=tab_1}, 2020.

\bibitem{american_diab_report}
A.~D. Association, ``Diabetes statistics by state.''
  \url{https://www.diabetes.org/resources/statistics/statistics-by-state},
  2020.

\bibitem{american2002screening}
A.~D. Association {\em et~al.}, ``Screening for diabetes,'' {\em Diabetes
  care}, vol.~25, no.~suppl 1, pp.~s21--s24, 2002.

\bibitem{diabetesRateReport}
U.~D. of~Health, ``National diabetes statistics report 2020.''
  \url{https://www.cdc.gov/diabetes/pdfs/data/statistics/national-diabetes-statistics-report.pdf},
  2020.

\bibitem{diab_types}
N.~I. of~Diabetes, Digestive, and K.~Diseases, ``What is diabetes?.''
  \url{https://www.niddk.nih.gov/health-information/diabetes/overview/what-is-diabetes/monogenic-neonatal-mellitus-mody},
  2020.

\bibitem{monogenic}
N.~I. of~Diabetes, Digestive, and K.~Diseases, ``Monogenic diabetes (neonatal
  diabetes mellitus and mody).''
  \url{https://www.niddk.nih.gov/health-information/diabetes/overview/what-is-diabetes/monogenic-neonatal-mellitus-mody},
  2020.

\bibitem{Cystic}
C.~F. Foundation, ``Cystic fibrosis-related diabetes.''
  \url{https://www.cff.org/Life-With-CF/Daily-Life/Cystic-Fibrosis-related-Diabetes/},
  2020.

\bibitem{atkinson2014type}
M.~A. Atkinson, G.~S. Eisenbarth, and A.~W. Michels, ``Type 1 diabetes,'' {\em
  The Lancet}, vol.~383, no.~9911, pp.~69--82, 2014.

\bibitem{devendra2004type}
D.~Devendra, E.~Liu, and G.~S. Eisenbarth, ``Type 1 diabetes: recent
  developments,'' {\em Bmj}, vol.~328, no.~7442, pp.~750--754, 2004.

\bibitem{cdc_diab1}
C.~for Disease~Control and Prevention, ``Type 1 diabetes.''
  \url{https://www.cdc.gov/diabetes/basics/type1.html}, 2020.

\bibitem{InToLife}
InToLife, ``Type 1 and 2 diabetes.''
  \url{https://www.intolife.in/about-diabetes/types-of-diabetes/type-1-diabetes},
  2020.

\bibitem{chatterjee2017type}
S.~Chatterjee, K.~Khunti, and M.~J. Davies, ``Type 2 diabetes,'' {\em The
  lancet}, vol.~389, no.~10085, pp.~2239--2251, 2017.

\bibitem{olokoba2012type}
A.~B. Olokoba, O.~A. Obateru, and L.~B. Olokoba, ``Type 2 diabetes mellitus: a
  review of current trends,'' {\em Oman medical journal}, vol.~27, no.~4,
  p.~269, 2012.

\bibitem{cdc_diab2}
C.~for Disease~Control and Prevention, ``Type 2 diabetes.''
  \url{https://www.cdc.gov/diabetes/basics/type2.html}, 2020.

\bibitem{gestationalDiabetes}
e.~a. E.~Filho, ``Heterogeneous methodology to support the early diagnosis of
  gestational diabetes,'' {\em IEEE Access}, vol.~7, pp.~67190--67199, 2019.

\bibitem{american2004gestational}
A.~D. Association {\em et~al.}, ``Gestational diabetes mellitus,'' {\em
  Diabetes care}, vol.~27, no.~suppl 1, pp.~s88--s90, 2004.

\bibitem{gdmOutcomes}
J.~L. P. E.~S. P.~M.~Rehder, B. G.~Pereira, ``Gestational and neonatal outcomes
  in women with positive screening for diabetes mellitus and 100g oral glucose
  challenge test normal,'' {\em Brazilian J. Gynecol. Obstetrics}, vol.~33,
  no.~2, pp.~81--86, 2010.

\bibitem{prediabetes}
J.~S. S.~Kacker, N.~Saboo, ``Prediabetes: Pathogenesis and adverse outcomes,''
  {\em International Journal of Medical Research Professionals}, vol.~4,
  pp.~1--6, March 2018.

\bibitem{steck2011review}
A.~K. Steck and W.~E. Winter, ``Review on monogenic diabetes,'' {\em Current
  Opinion in Endocrinology, Diabetes and Obesity}, vol.~18, no.~4,
  pp.~252--258, 2011.

\bibitem{monogenicDiabetes}
e.~a. Carmody~D, {\em A clinical guide to monogenic diabetes, Genetic Diagnosis
  of Endocrine Disorders}.
\newblock Elsevier, 2~ed., 2016.

\bibitem{shields2010maturity}
B.~Shields, S.~Hicks, M.~Shepherd, K.~Colclough, A.~T. Hattersley, and
  S.~Ellard, ``Maturity-onset diabetes of the young (mody): how many cases are
  we missing?,'' {\em Diabetologia}, vol.~53, no.~12, pp.~2504--2508, 2010.

\bibitem{monogenicGeneticTest}
NIH, ``Monogenic diabetes (neonatal diabetes mellitus \& mody).''
  \url{https://www.niddk.nih.gov/health-information/diabetes/overview/what-is-diabetes/monogenic-neonatal-mellitus-mody.html},
  2020.

\bibitem{moran2009cystic}
A.~Moran, J.~Dunitz, B.~Nathan, A.~Saeed, B.~Holme, and W.~Thomas, ``Cystic
  fibrosis--related diabetes: current trends in prevalence, incidence, and
  mortality,'' {\em Diabetes care}, vol.~32, no.~9, pp.~1626--1631, 2009.

\bibitem{CFRDStatistics}
M.~H. Kayani~K, Mohammed~R, ``Cystic fibrosis-related diabetes,'' {\em
  Frontiers in endocrinology}, vol.~9, February 2018.

\bibitem{CFRDTreatment}
N.~Bridges, R.~Rowe, and R.~I.~G. Holt, ``Unique challenges of cystic
  fibrosis-related diabetes,'' {\em Diabetic Medicine}, vol.~35, no.~9,
  pp.~1181--1188, 2018.

\bibitem{cook2008ethics}
A.~F. Cook and H.~Hoas, ``Ethics and rural healthcare: What really happens?
  what might help?,'' {\em The American Journal of Bioethics}, vol.~8, no.~4,
  pp.~52--56, 2008.

\bibitem{merwin2006differential}
E.~Merwin, A.~Snyder, and E.~Katz, ``Differential access to quality rural
  healthcare: professional and policy challenges,'' {\em Family and Community
  Health}, pp.~186--194, 2006.

\bibitem{maan2016survey}
A.~K. Maan, D.~A. Jayadevi, and A.~P. James, ``A survey of memristive threshold
  logic circuits,'' {\em IEEE transactions on neural networks and learning
  systems}, vol.~28, no.~8, pp.~1734--1746, 2016.

\bibitem{huang2006extreme}
G.-B. Huang, Q.-Y. Zhu, and C.-K. Siew, ``Extreme learning machine: theory and
  applications,'' {\em Neurocomputing}, vol.~70, no.~1-3, pp.~489--501, 2006.

\bibitem{matias2014learning}
T.~Matias, F.~Souza, R.~Ara{\'u}jo, and C.~H. Antunes, ``Learning of a
  single-hidden layer feedforward neural network using an optimized extreme
  learning machine,'' {\em Neurocomputing}, vol.~129, pp.~428--436, 2014.

\bibitem{hornik1991approximation}
K.~Hornik, ``Approximation capabilities of multilayer feedforward networks,''
  {\em Neural networks}, vol.~4, no.~2, pp.~251--257, 1991.

\bibitem{huang1998upper}
G.-B. Huang and H.~A. Babri, ``Upper bounds on the number of hidden neurons in
  feedforward networks with arbitrary bounded nonlinear activation functions,''
  {\em IEEE transactions on neural networks}, vol.~9, no.~1, pp.~224--229,
  1998.

\bibitem{huang2000classification}
G.-B. Huang, Y.-Q. Chen, and H.~A. Babri, ``Classification ability of single
  hidden layer feedforward neural networks,'' {\em IEEE Transactions on Neural
  Networks}, vol.~11, no.~3, pp.~799--801, 2000.

\bibitem{zheng2013text}
W.~Zheng, Y.~Qian, and H.~Lu, ``Text categorization based on regularization
  extreme learning machine,'' {\em Neural Computing and Applications}, vol.~22,
  no.~3, pp.~447--456, 2013.

\bibitem{karpagachelvi2012classification}
S.~Karpagachelvi, M.~Arthanari, and M.~Sivakumar, ``Classification of
  electrocardiogram signals with support vector machines and extreme learning
  machine,'' {\em Neural Computing and Applications}, vol.~21, no.~6,
  pp.~1331--1339, 2012.

\bibitem{elsayed2020simple}
N.~Elsayed and Z.~S. Zaghloul, ``A simple extreme learning machine model for
  detecting heart arrhythmia in the electrocardiography signal,'' in {\em 2020
  IEEE 63rd International Midwest Symposium on Circuits and Systems (MWSCAS)},
  pp.~513--516, IEEE, 2020.

\bibitem{pan2012leukocyte}
C.~Pan, D.~S. Park, Y.~Yang, and H.~M. Yoo, ``Leukocyte image segmentation by
  visual attention and extreme learning machine,'' {\em Neural Computing and
  Applications}, vol.~21, no.~6, pp.~1217--1227, 2012.

\bibitem{feng2012evolutionary}
G.~Feng, Z.~Qian, and X.~Zhang, ``Evolutionary selection extreme learning
  machine optimization for regression,'' {\em Soft Computing}, vol.~16, no.~9,
  pp.~1485--1491, 2012.

\bibitem{bartlett1998thesamplecomplexityofp}
P.~Bartlett, ``Thesamplecomplexityofp atternclassification withneuralnetworks:
  Thesizeoftheweightsismo reimportantthan thesizeofthenetwork,'' {\em
  IEEETrans. Inf. Theory}, vol.~44, no.~2, 1998.

\bibitem{huang2015extreme}
G.-B. Huang, ``What are extreme learning machines? filling the gap between
  frank rosenblatt’s dream and john von neumann’s puzzle,'' {\em Cognitive
  Computation}, vol.~7, no.~3, pp.~263--278, 2015.

\bibitem{baumann2017co2}
M.~Baumann, J.~Peters, M.~Weil, and A.~Grunwald, ``Co2 footprint and life-cycle
  costs of electrochemical energy storage for stationary grid applications,''
  {\em Energy Technology}, vol.~5, no.~7, pp.~1071--1083, 2017.

\bibitem{k01r-x481-21}
K.~Kanagarathinam, ``Early stage diabetes risk prediction dataset,'' 2021.

\bibitem{misc_early_stage_diabetes_risk_prediction_dataset._529}
``{Early stage diabetes risk prediction dataset.}.'' UCI Machine Learning
  Repository, 2020.

\bibitem{pottman1992identification}
M.~Pottman and D.~E. Seborg, ``Identification of non-linear processes using
  reciprocal multiquadric functions,'' {\em Journal of Process Control},
  vol.~2, no.~4, pp.~189--203, 1992.

\end{thebibliography}
\bibliographystyle{ieeetr}

\end{document}